\documentclass[10pt,twocolumn,letterpaper]{article}

\usepackage[pagenumbers]{cvpr} 

\usepackage{graphicx}
\usepackage{amsmath}
\usepackage{amssymb}
\usepackage{booktabs}
\usepackage{paralist}

\usepackage[pagebackref,breaklinks,colorlinks]{hyperref}

\usepackage[capitalize]{cleveref}
\crefname{section}{Sec.}{Secs.}
\Crefname{section}{Section}{Sections}
\Crefname{table}{Table}{Tables}
\crefname{table}{Tab.}{Tabs.}

\def\cvprPaperID{*****} 
\def\confName{AI535}
\def\confYear{S2022}

\begin{document}

\title{Convolutional Deep Colorization for Image Compression: A Color Grid Based Approach}

\author{Ian Tassin \hspace{20pt} Kristen Goebel\hspace{20 pt} Brittany Lasher\\
Oregon State University\\
{\tt\small \{tassini, goebelk, lasherb\}@oregonstate.edu}
}
\maketitle
\begin{abstract}
    The search for image compression optimization techniques is a topic of constant interest both in and out of academic circles. One method that shows promise toward future improvements in this field is image colorization since image colorization algorithms can reduce the amount of color data that needs to be stored for an image. Our work focuses on optimizing a `color grid' based approach to fully-automated image color information retention with regard to convolutional colorization network architecture for the purposes of image compression. More generally, using a convolutional neural network for image re-colorization, we want to minimize the amount of color information that is stored while still being able to faithfully re-color images. Our results yielded a promising image compression ratio, while still allowing for successful image recolorization reaching high CSIM values.
\end{abstract}

\tableofcontents
\newpage

\section{Research Overview}
\label{sec:intro}
Colorization of decolorized or black and white images is an important task that can be completed through deep learning approaches. Previous studies have identified that retaining points of color within an image lead to improved results when coloring images. In this work, we aim to investigate how retaining a different number of colored pixels within an image will help with reconstitution of the original colors, when using a convolutional neural network. We also evaluate how partial decolorization of images improves compression or storage size. 

\subsection{Introduction}
    Image colorization of grey-scale images to increase visual appeal is a research area with over 20 years of history \cite{ZegerIvana2021GICM}. The primary appeal of this research historically has focused on improving images aesthetics by the addition of color. Particularly this has been motivated by a desire to breathe new life into historical photographs taken before color photography.

    A subsequent off shoot of this research has considered the potential for image decolorization as a method of image compression since it requires less data to store a grey-scale image compared to a color image. Given a sufficiently accurate model for re-colorization of the image when it needs to be retrieved this could lead to image storage improvements. Retaining some color information has been shown to lead to dramatic improvements in colorization \cite{XiaoYi2022IDCa}. However, without user intervention, knowing at what locations to store color information proves tricky and, in research thus far, necessitates the storage of more color information in a systematic way. These automated methods have also proven very effective \cite{FatimaAroosh2021Gitn}.

\subsection{Related Work}
    With the rise of deep learning, many different studies have focused on a broad range of deep learning techniques and model architectures to accomplish the task of adding color to black and white images \cite{Cheng2015}. These approaches can quickly and easily handle large amounts of data and have demonstrated effective results. Since the first deep learning image colorization method in 2015 \cite{Cheng2015}, the field has rapidly grown. Various approaches have been applied, such as convolutional neural networks \cite{Larsson2016, He2018, Zhang2017, Su2020, Dong2022}, which allow for retaining spatial information, generative adversarial networks (GANs) \cite{Zhang2018, Kuang2020}, which are advantageous for preserving details and high-quality image generation, and transformer networks \cite{Kumar2021} that excel at applying color with context to the entire image. Previous studies have also identified that color fidelity was improved when some original color from the image was retained \cite{FatimaAroosh2021Gitn, Boutarfass2020}. In \cite{FatimaAroosh2021Gitn}, they applied two different automated approaches for choosing the pixel location(s) for color retention. Such approaches included grid-based, where every n-th pixel remained colored, and segment-based, where segments in the images were identified and a pixel within each segment was chosen to retain color. Another recent study of particular interest for our paper is \cite{XiaoYi2022IDCa} which proposes a convolutional, U-net architecture for image recolorization. Hence forth we will refer to this model architecture as \textbf{XiaoNet}. Our study will focus on the marriage of two methods: the use of grid-based, automated color retention like the method described in \cite{FatimaAroosh2021Gitn} with a convolutional U-net architecture derived from the XiaoNet architecture. Furthermore, we aim to find the optimal amount of color-information retention to maximize the image compression while maintaining enough information to make highly accurate predictions using our XiaoNet-like architecture More information on our architecture is available in Section \ref{sec:Model Arch}. 

\section{Methodology}

\begin{figure}[t]
    \centering
    \fbox{\rule{0pt}{0.1in} \includegraphics[scale=0.15]{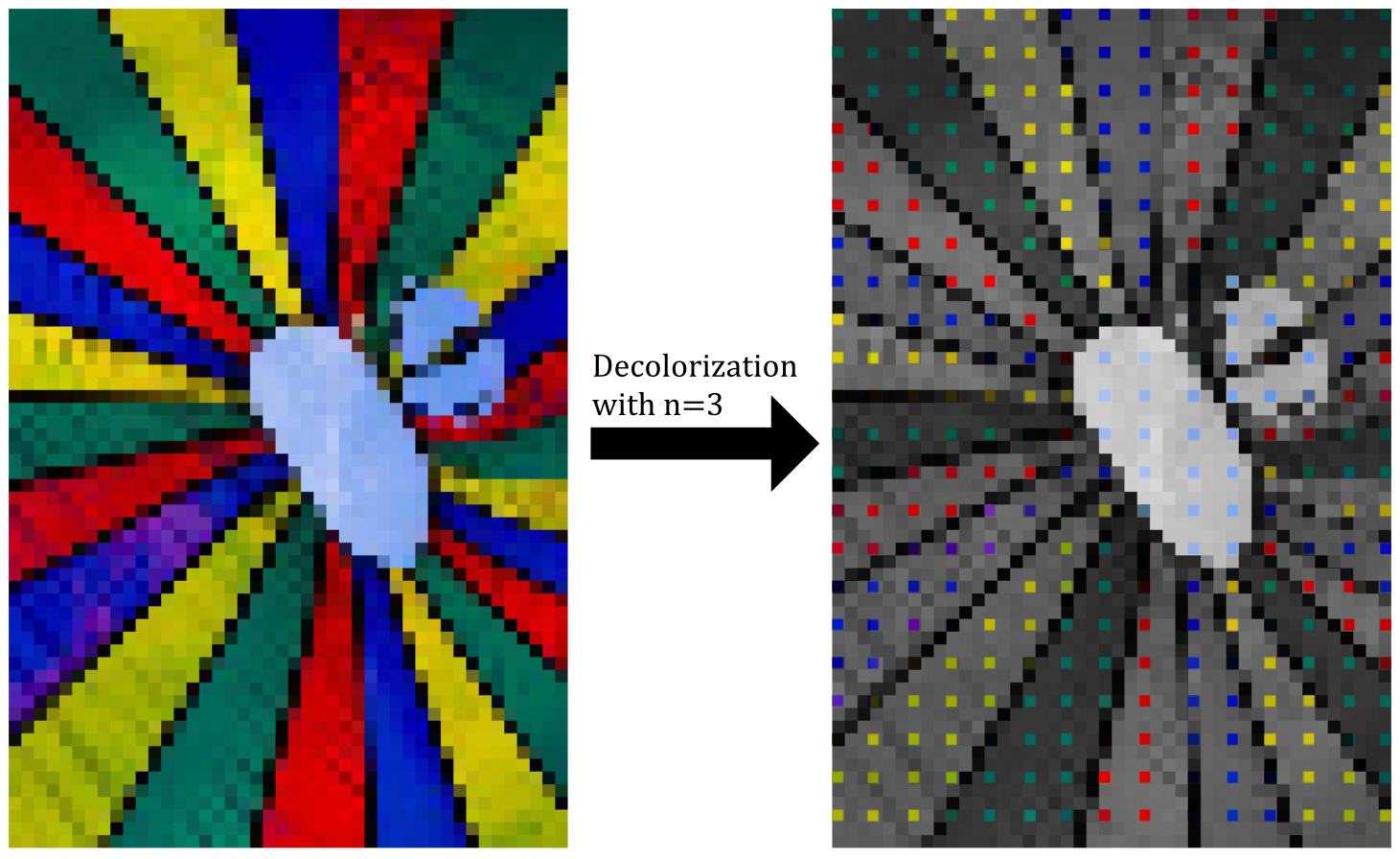} \rule{0pt}{0.1in}}
    \caption{Example decolorization process on low resolution image with n=3.}
    \label{fig:decolorization grid}
    \end{figure}

    We used the Crayon model architecture, see Section \ref{sec:Model Arch}, and passed it color information at regular intervals along a `color grid' in the AB color channels. For an example of such a color grid for \textit{n}=3 see Figure \ref{fig:decolorization grid}. A similar, grid-based color encoding was used in \cite{FatimaAroosh2021Gitn}, however they use a significantly different network architecture. Our paper uses a grid-based color encoding with a model architecture derived from \cite{XiaoYi2022IDCa}.
    
    \subsection{Data Set} Within this work, we analyzed the results of our method on the Imagenette version two data set, which is composed of 9469 training images and 3925 validation images. We removed 50 images from the validation set to use as a test set for our experiments. For consistency and convenience, we cropped all images to size 320x320 regardless of \textit{n} value before inputting them into the model. 
    
    \subsection{Data Processing} In order to test our model on colorization, we needed to remove the the bulk of color from the images in our data set. To accomplish this task, we first converted the color from RGB color space to LAB color space as is standard for image colorization tasks. The LAB color mode represents greyscale as a single channel lightness (L) and represents color information with 2 channels (A and B). This color space in generally believed to better match human perception of color, and increase the semantic meaning of distance measurements. Next, most of the color was removed from each image in the data set, retaining only every \textit{n}th colored pixel in a grid formation. This will enable a smaller storage size of images, while also retaining key color information advantageous for color regeneration. In this work, we test \textit{n}-values of 6, 15, 20, 40, 50, 60, 80, and 100, to determine what level of retained color results in the highest overall performance.

    \subsection{Training}

    \subsubsection{Loss function} We used mean squared error (MSE) as our loss function, rather than the Huber loss used in \cite{XiaoYi2022IDCa}, based on empirically better results when training our model. 

    \subsubsection{Optimizer} For model training we used the ADAM optimizer with a step size of $10^{-4}$.

    \subsubsection{Training Evaluation}
    We collected data across many color-grid spacing values, \textit{n}. For each value of \textit{n}, we trained a model for 30 epochs. After each epoch we saved our model if its validation accuracy was higher than the previous version, these best-forming trained models are heretoafter referred to as the \textbf{canonical model} for a given \textit{n} value.

    Once training was complete for each \textit{n} value we evaluated the canonical model for that \textit{n} value on our test set. Our evaluation metrics were calculated with regard to color-tone similarity index measure (CSIM) and peak signal to noise ratio (PSNR). These results, along with examples of the models predictions for each \textit{n} value can be examined in more detail in Section \ref{sec:Eval}.

    \section{The Crayon Model Architecture}
    \label{sec:Model Arch}

    \subsection{Overview}
    \label{sec:Model Arch overview}
    Our model architecture, \textbf{Crayon}, which stands for \textbf{C}onvolutional \textbf{R}ecolorization \textbf{A}rchitecture for \textbf{Y}ielding \textbf{O}cular \textbf{N}iceties, was derived from the XiaoNet model architecture \cite{XiaoYi2022IDCa}. Crayon consists of a 4 stage U-net followed by a small residual network. A detailed breakdown of the model can be found in Section \ref{sec:ModArch Details}.

    \subsection{Architecture Details}
    \label{sec:ModArch Details}
    A full breakdown of the Crayon architecture can be found in tables \ref{tab:model-arch} and \ref{tab:model-arch-pt2}.\\
    
    \textbf{Key for Tables \ref{tab:model-arch} and \ref{tab:model-arch-pt2}.}
    \begin{itemize}
        \item $X_i$ indicates spacial dimensions (width and height) of the input.
        \item $X_o$ indicates spacial dimensions (width and height) of the output.
        \item $C_i$ indicates the number of channels of input.
        \item $C_o$ indicates the number of channels of output.
        \item $K$ indicates the kernel size. 
        \item $S$ indicates the stride.
        \item $P$ indicates the padding. 
        \item $D$ indicates the dilation.
        \item ``Data From" indicates which layer(s) output is input to the current layer. \begin{itemize}
            \item $[x,y]$ Brackets indicate that the output of the contained layers is being concatenated together along the channel axis.
            \item $\{x, y\}$ Curly brackets braces indicates the contained layers are undergoing element wise addition.
        \end{itemize}
    
    \end{itemize}
\begin{table*}[t]
    \begin{center}
    \begin{tabular}{||c c c c c c c c c c c||} 
     \hline
     Layer Index & Operation & $X_i$ & $X_o$ & $C_i$ & $C_o$ & $K$ & $S$ & $P$ & $D$ & Data From \\ [0.5ex] 
     \hline\hline
     0 & L Channel Input & N/A & $320$ & N/A & $1$ & N/A & N/A & N/A & N/A & N/A \\
     1 & AB Channel Input & N/A & $320$ & N/A & $2$ & N/A & N/A & N/A & N/A & N/A \\
     \hline
     2 & Convolution & $320$ & $320$ & $3$ & $64$ & $3$ & $1$ & $1$ & $1$ & $[0,1]$ \\
     3 & ReLU & $320$ & $320$ & $64$ & $64$ & N/A & N/A & N/A & N/A & $2$ \\
     4 & Convolution & $320$ & $320$ & $64$ & $64$ & $3$ & $1$ & $1$ & $1$ & $3$ \\
     5 & ReLU & $320$ & $320$ & $64$ & $64$ & N/A & N/A & N/A & N/A & $4$ \\
     6 & MaxPool & $320$ & $160$ & $64$ & $64$ & $2$ & $2$ & $0$ & $1$ & $5$ \\
     \hline
     7 & Convolution & $160$ & $160$ & $64$ & $128$ & $3$ & $1$ & $1$ & $1$ & $6$ \\
     8 & ReLU & $160$ & $160$ & $128$ & $128$ & N/A & N/A & N/A & N/A & $7$ \\
     9 & Convolution & $160$ & $160$ & $128$ & $128$ & $3$ & $1$ & $1$ & $1$ & $8$ \\
     10 & ReLU & $160$ & $160$ & $128$ & $128$ & N/A & N/A & N/A & N/A & $9$ \\
     11 & MaxPool & $160$ & $80$ & $128$ & $128$ & $2$ & $2$ & $0$ & $1$ & $10$ \\
     \hline
     12 & Convolution & $80$ & $80$ & $128$ & $256$ & $3$ & $1$ & $1$ & $1$ & $11$ \\
     13 & ReLU & $80$ & $80$ & $256$ & $256$ & N/A & N/A & N/A & N/A & $12$ \\
     14 & Convolution & $80$ & $80$ & $256$ & $256$ & $3$ & $1$ & $1$ & $1$ & $13$ \\
     15 & ReLU & $80$ & $80$ & $256$ & $256$ & N/A & N/A & N/A & N/A & $14$ \\
     16 & MaxPool & $80$ & $40$ & $256$ & $256$ & $2$ & $2$ & $0$ & $1$ & $15$ \\
     \hline
     17 & Convolution & $40$ & $40$ & $256$ & $512$ & $3$ & $1$ & $1$ & $1$ & $16$ \\
     18 & ReLU & $40$ & $40$ & $512$ & $512$ & N/A & N/A & N/A & N/A & $17$ \\
     19 & Convolution & $40$ & $40$ & $512$ & $512$ & $3$ & $1$ & $1$ & $1$ & $18$ \\
     20 & ReLU & $40$ & $40$ & $512$ & $512$ & N/A & N/A & N/A & N/A & $19$ \\
     \hline
     21 & Convolution & $40$ & $40$ & $512$ & $512$ & $3$ &  $1$ & $2$ & $2$ & $20$ \\
     22 & ReLU & $40$ & $40$ & $512$ & $512$ & N/A & N/A & N/A & N/A & $21$ \\
     23 & Convolution & $40$ & $40$ & $512$ & $512$ & $3$ & $1$ & $2$  & $2$ & $22$ \\
     24 & ReLU & $40$ & $40$ & $512$ & $512$ & N/A & N/A & N/A & N/A & $23$ \\
     25 & Convolution & $40$ & $40$ & $512$ & $512$ & $3$ & $1$ & $2$ & $2$ & $24$ \\
     26 & ReLU & $40$ & $40$ & $512$ & $512$ & N/A & N/A & N/A & N/A & $25$ \\
     27 & Convolution & $40$ & $40$ & $512$ & $512$ & $3$ & $1$ & $2$ & $2$ & $26$ \\
     28 & ReLU & $40$ & $40$ & $512$ & $512$ & N/A & N/A & N/A & N/A & $27$ \\
     29 & Convolution & $40$ & $40$ & $512$ & $512$ & $3$ & $1$ & $2$ & $2$ & $28$ \\
     30 & ReLU & $40$ & $40$ & $512$ & $512$ & N/A & N/A & N/A & N/A & $29$ \\
     31 & Convolution & $40$ & $40$ & $512$ & $512$ & $3$ & $1$ & $2$ & $2$ & $30$ \\
     32 & ReLU & $40$ & $40$ & $512$ & $512$ & N/A & N/A & N/A & N/A & $31$ \\
     \hline
     33 & Convolution & $40$ & $40$ & $512$ & $512$ & $3$ & $1$ & $1$ & $1$ & $32$ \\
     34 & ReLU & $40$ & $40$ & $512$ & $512$ & N/A & N/A & N/A & N/A & $33$ \\
     \hline
     35 & Transposed Conv. & $40$ & $80$ & $512$ & $256$ & $2$ & $2$ & $0$ & $1$ & $34$ \\
     36 & ReLU & $80$ & $80$ & $256$ & $256$ & N/A & N/A & N/A & N/A & $35$ \\
     37 & Concatenation & $80$ & $80$ & $256+256$ & $512$ & N/A & N/A & N/A & N/A & $[15,36]$ \\
     38 & Convolution & $80$ & $80$ & $512$ & $256$ & $3$ & $1$ & $1$ & $1$ & $37$ \\
     39 & ReLU & $80$ & $80$ & $256$ & $256$ & N/A & N/A & N/A & N/A & $38$ \\
     \hline
     40 & Transposed Conv. & $80$ & $160$ & $256$ & $128$ & $2$ & $2$ & $0$ & $1$ & $39$ \\
     41 & ReLU & $160$ & $160$ & $128$ & $128$ & N/A & N/A & N/A & N/A & $40$ \\
     42 & Concatenation & $160$ & $160$ & $128+128$ & $256$ & N/A & N/A & N/A & N/A & $[10,41]$ \\
     43 & Convolution & $160$ & $160$ & $256$ & $128$ & $3$ & $1$ & $1$ & $1$ & $42$ \\
     44 & ReLU & $160$ & $160$ & $128$ & $128$ & N/A & N/A & N/A & N/A & $43$ \\
     \hline
     45 & Transposed Conv. & $160$ & $320$ & $128$ & $128$ & $2$ & $2$ & $0$ & $1$ & $44$ \\
     46 & ReLU & $320$ & $320$ & $128$ & $128$ & N/A & N/A & N/A & N/A & $45$ \\
     47 & Concatenation & $320$ & $320$ & $64+128$ & $192$ & N/A & N/A & N/A & N/A & $[5,46]$ \\
     48 & Convolution & $320$ & $320$ & $192$ & $128$ & $3$ & $1$ & $1$ & $1$ & $47$ \\
     49 & ReLU & $320$ & $320$ & $128$ & $128$ & N/A & N/A & N/A & N/A & $48$ \\
     50 & Convolution & $320$ & $320$ & $128$ & $2$ & $1$ & $1$ & $0$ & $1$ & $49$\\
     ... & \textbf{Continued} & \textbf{on} & \textbf{table} & \ref{tab:model-arch-pt2} & ... & ... & ... & ... & ... & ...\\
     \hline
    \end{tabular}
    \caption{Crayon Model Architecture (Part 1): U-Net}
    \label{tab:model-arch}
    \end{center}
\end{table*}

\begin{table*}[t]
    \begin{center}
    \begin{tabular}{||c c c c c c c c c c c||} 
     \hline
     Layer Index & Operation & $X_i$ & $X_o$ & $C_i$ & $C_o$ & $K$ & $S$ & $P$ & $D$ & Data From \\ [0.5ex] 
     \hline\hline
     51 & Concatenation & $320$ & $320$ & $1+2+2$ & $3$ & N/A & N/A & N/A & N/A & $[0,1,50]$ \\
     \hline
     52 & Convolution & $320$ & $320$ & $5$ & $64$ & $3$ & $1$ & $1$ & $1$ & $51$ \\
     53 & ReLU & $320$ & $320$ & $64$ & $64$ & N/A & N/A & N/A & N/A & $52$ \\
     54 & Convolution & $320$ & $320$ & $64$ & $64$ & $3$ & $1$ & $1$ & $1$ & $53$ \\
     55 & ReLU & $320$ & $320$ & $64$ & $64$ & N/A & N/A & N/A & N/A & $54$ \\
     \hline
     56 & Convolution & $320$ & $320$ & $64$ & $64$ & $3$ & $1$ & $1$ & $1$ & $55$ \\
     57 & ReLU & $320$ & $320$ & $64$ & $64$ & N/A & N/A & N/A & N/A & $56$ \\
     58 & Convolution & $320$ & $320$ & $64$ & $64$ & $3$ & $1$ & $1$ & $1$ & $57$ \\
     59 & ReLU & $320$ & $320$ & $64$ & $64$ & N/A & N/A & N/A & N/A & $58$ \\
     60 & Addition & $320+320$ & $320$ & $64+64$ & $64$ & N/A & N/A & N/A & N/A & $\{54, 59\}$ \\
     \hline
     61 & Convolution & $320$ & $320$ & $64$ & $64$ & $3$ & $1$ & $1$ & $1$ & $60$ \\
     62 & ReLU & $320$ & $320$ & $64$ & $64$ & N/A & N/A & N/A & N/A & $61$ \\
     63 & Convolution & $320$ & $320$ & $64$ & $64$ & $3$ & $1$ & $1$ & $1$ & $62$ \\
     64 & ReLU & $320$ & $320$ & $64$ & $64$ & N/A & N/A & N/A & N/A & $63$ \\
     65 & Addition & $320+320$ & $320$ & $64+64$ & $64$ & N/A & N/A & N/A & N/A & $\{60, 64\}$ \\
     \hline
     66 & Convolution & $320$ & $320$ & $64$ & $64$ & $3$ & $1$ & $1$ & $1$ & $65$ \\
     67 & ReLU & $320$ & $320$ & $64$ & $64$ & N/A & N/A & N/A & N/A & $66$ \\
     68 & Convolution & $320$ & $320$ & $64$ & $64$ & $3$ & $1$ & $1$ & $1$ & $67$ \\
     69 & ReLU & $320$ & $320$ & $64$ & $64$ & N/A & N/A & N/A & N/A & $68$ \\
     70 & Addition & $320+320$ & $320$ & $64+64$ & $64$ & N/A & N/A & N/A & N/A & $\{65, 69\}$ \\
     \hline
     71 & Convolution & $320$ & $320$ & $64$ & $64$ & $3$ & $1$ & $1$ & $1$ & $70$ \\
     72 & ReLU & $320$ & $320$ & $64$ & $64$ & N/A & N/A & N/A & N/A & $71$ \\
     73 & Convolution & $320$ & $320$ & $64$ & $64$ & $3$ & $1$ & $1$ & $1$ & $72$ \\
     74 & ReLU & $320$ & $320$ & $64$ & $64$ & N/A & N/A & N/A & N/A & $73$ \\
     75 & Addition & $320+320$ & $320$ & $64+64$ & $64$ & N/A & N/A & N/A & N/A & $\{70, 74\}$ \\
     \hline
     76 & Convolution & $320$ & $320$ & $64$ & $256$ & $3$ & $1$ & $1$ & $1$ & $75$ \\
     77 & ReLU & $320$ & $320$ & $64$ & $64$ & N/A & N/A & N/A & N/A & $76$ \\
     78 & Convolution & $320$ & $320$ & $256$ & $2$ & $3$ & $1$ & $1$ & $1$ & $77$ \\
     79 & Concatenation & $320$ & $320$ & $1+2$ & $3$ & N/A & N/A & N/A & N/A & $[0,78]$ \\
     \hline
    \end{tabular}
    \caption{Crayon Model Architecture (Part 2): Residual Network}
    \label{tab:model-arch-pt2}
    \end{center}
\end{table*}

\section{Evaluation of Colorization}
\label{sec:Eval}
To evaluate our image colorization performance, we examine two metrics, color-tone similarity index measure (CSIM) and peak signal to noise ratio (PSNR).

    \subsection{Results and Data}
    Each model is evaluated on a held-out test set of 50 images. The recolorization quality of the images is evaluted using CSIM and PSNR. For both metrics, a higher value indicated a better quality recoloring. The quality of the images for each \textit{n} as well as a theoretical upper bound for the compression size relative to the original image size are shown in Figure \ref{fig:eval_plot}. As \textit{n} increases and retained color information becomes sparser, both PSNR and CSIM decrease.

    We calculate the theoretical upper bound for the image compression size, relative to the original image, using  \begin{equation}
      relativeSize = \frac{1}{3} + \frac{1}{n^2}
      \label{eq:im-size}
    \end{equation}
    which is a combination of the size of the grey scale image and the remaining colored pixels. As n increases, the overall size will approach 1/3, so any compression size near this value is maximizing the capabilities of this method. The best \textit{n} appears to be around \textit{n}=20, after this point the compression size does not improve significantly but the image recolorization continues to decrease in quality. 
    
    The differences in recolorization on an example test image are shown in Figure \ref{fig:recolor_grid}. After \textit{n}=20, the differences between the recolorization and the original image are more easily noticed, such as a lack of red in the window frames or a brown coloring on the walls of the building.
    The CSIM results we observed for lower n-values ($<20$) are comparable to results observed through a GAN-based colorization approach, where the CSIM of their autocolorization mode reached values of roughly 0.87 - 0.88. When comparing our PSNR results to other method's results (\cref{fig:eval_plot}) \cite{Baldassarre2017, Jwa2021, Zhang2016}, our method reached a higher PSNR. However, this is not a direct comparison, as the data set and number of samples varies between our evaluation and other studies. Furthermore, when examining our compression results to those from other studies, we found them to be comparable \cite{6236598}.

    \subsection{Conclusion}
\begin{figure}[t]
    \centering
    \fbox{\rule{0pt}{0.1in} \includegraphics[scale=0.45]{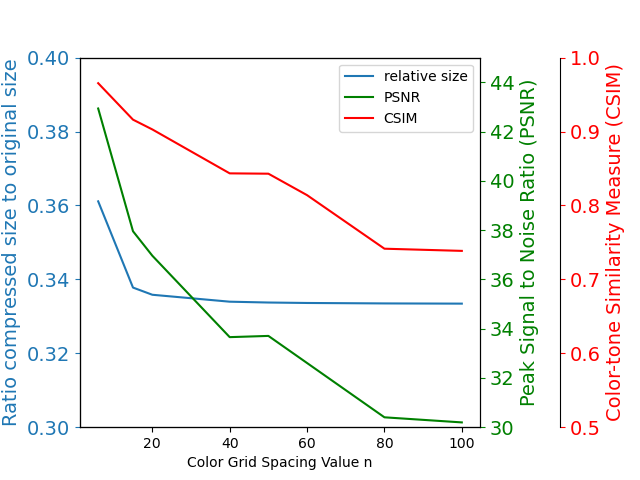} \rule{0pt}{0.1in}}
    \caption{Image quality and relative compression size of images for varied color grid spacing values.}
    \label{fig:eval_plot}
    \end{figure}

\begin{figure}[t]
\centering
\begin{tabular}{ccc}
\subcaptionbox{n=6\label{n6}}{\includegraphics[width = 0.9in]{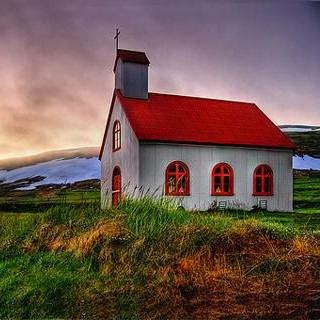}} &
\subcaptionbox{n=15\label{n15}}{\includegraphics[width = 0.9in]{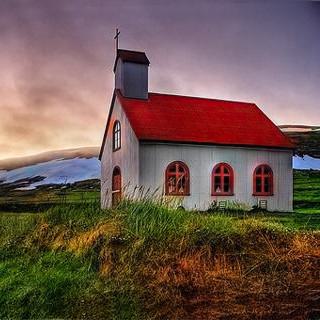}} &
\subcaptionbox{n=20\label{n20}}{\includegraphics[width = 0.9in]{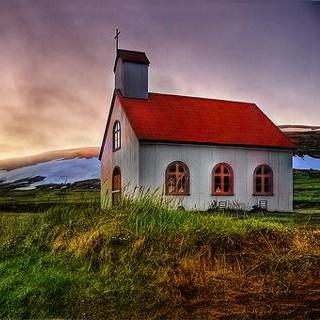}}\\
\subcaptionbox{n=40\label{n40}}{\includegraphics[width = 0.9in]{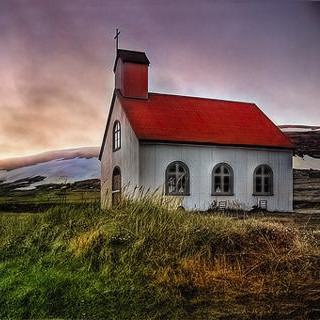}} &
\subcaptionbox{original\label{gt}}{\includegraphics[width = 0.9in]{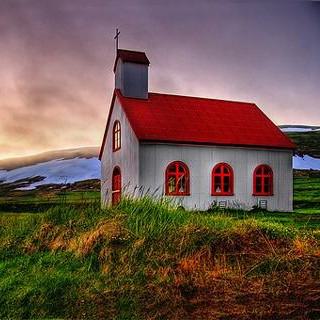}} &
\subcaptionbox{n=50\label{n50}}{\includegraphics[width = 0.9in]{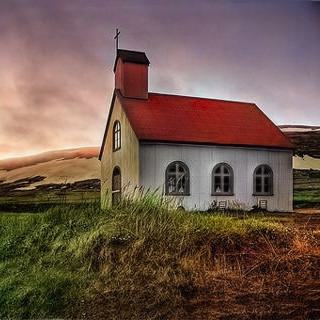}}\\
\subcaptionbox{n=60\label{n60}}{\includegraphics[width = 0.9in]{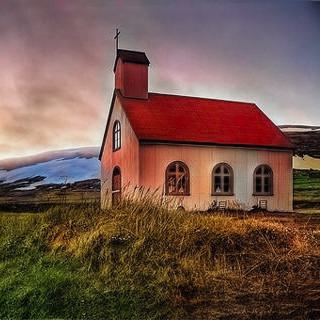}} &
\subcaptionbox{n=80\label{n80}}{\includegraphics[width = 0.9in]{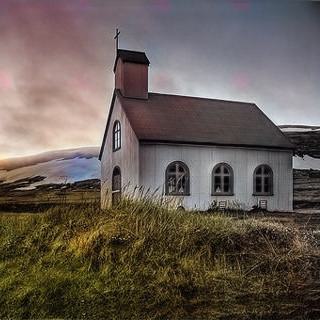}} &
\subcaptionbox{n=100\label{n100}}{\includegraphics[width = 0.9in]{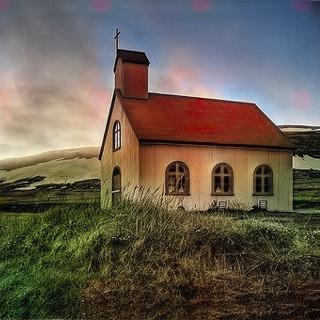}}
\end{tabular}
\caption{Image recolorization with varying levels of color retention. As n increases, images look less like the original image.}
\label{fig:recolor_grid}
\end{figure}

    Our results are very promising regarding the used of recolorizaton as a form of image compression. We find very minimal losses with regard to PSNR and CSIM when using \textit{n} values between 6 and 15 (\cref{fig:eval_plot}) as well as little to no visual artifacts in the recolored images (\cref{fig:recolor_grid}). Additionally, the compression gain from values larger than $n$ = 20 are very minimal (\cref{eq:im-size}) while the number of visual artifacts (\cref{fig:recolor_grid}) and measured losses increase dramatically (\cref{fig:eval_plot}).

    It is not clear to what extent these results will generalize to all other model architectures, however a previous study using a GAN-based method for image colorization applied a similar approach to automatically choosing colored pixels, and showed successful results \cite{FatimaAroosh2021Gitn}. Additionally, alternative algorithmic approaches for selecting which pixels to keep or discard color information for may lead to improved performance. For example, future research could select color pixels based on their dis-similarity to adjacent pixels, this could ensure that small features are correctly colored instead of accidentally being skipped-over and having no encoded color information saved which can be a problem for this approach, especially for larger \textit{n} values.

    We doubt that encoding less $n$ = 20 color information will proved viable for any model architecture since the compression benefits are very minor above $n$ = 20 and the sparsity of color data above that range makes inaccurate coloring predictions much more likely.

{\small
\bibliographystyle{ieee_fullname}

}

\end{document}